\documentclass[conference]{IEEEtran}
\IEEEoverridecommandlockouts
\usepackage{cite}
\usepackage{amsmath,amssymb,amsfonts}
\usepackage{algorithmic}
\usepackage{graphicx}
\usepackage{textcomp}
\usepackage{xcolor}
\def\BibTeX{{\rm B\kern-.05em{\sc i\kern-.025em b}\kern-.08em
    T\kern-.1667em\lower.7ex\hbox{E}\kern-.125emX}}

\usepackage{ifpdf}
\usepackage{xcolor}
\usepackage{cite}
\usepackage{lipsum}
\usepackage{amsmath}
\usepackage{siunitx}
\usepackage{array}
\usepackage{graphicx}
\usepackage{url}
\usepackage{tabularx}
\usepackage[english]{babel}
\usepackage{algorithm}
\usepackage{algorithmic}
\usepackage{multirow}
\usepackage{hhline}
\usepackage{amsmath,amssymb}
\usepackage{multirow}
\usepackage{caption}
\usepackage{amssymb}
\usepackage{graphicx}
\usepackage{epstopdf}
\usepackage{subcaption}
\usepackage{cite}
\usepackage{siunitx}
\usepackage{color}
\usepackage{amssymb}

\usepackage{amsthm}
\usepackage{tikz}
\usepackage{mathtools}
\usepackage{siunitx}
\usepackage{listings}
\usepackage{eurosym}
\usepackage{footnote}
\usepackage{booktabs,caption}
\usepackage{pifont}

\usepackage{amsmath}

\DeclareSIUnit[inter-unit-product={}] \MVA {\mega\volt\ampere} %apparent power
\DeclareSIUnit[inter-unit-product={}] \MWh {\MW\hour} %apparent power
\DeclareSIUnit \pu {p.u.}
\DeclareSIUnit[inter-unit-product={}] \KV {\kilo\volt} %KV

\def\BibTeX{{\rm B\kern-.05em{\sc i\kern-.025em b}\kern-.08em
    T\kern-.1667em\lower.7ex\hbox{E}\kern-.125emX}}
\begin{document}

\title{A Comparative Study: Adaptive Fuzzy Inference Systems for Energy Prediction in Urban Buildings
}
\author{\IEEEauthorblockN{Mainak Dan}
\IEEEauthorblockA{\textit{Interdisciplinary Graduate School} \\
\textit{Nanyang Technological University}\\
Singapore \\
mainak001@e.ntu.edu.sg}
\and
\IEEEauthorblockN{Seshadhri Srinivasan}
\IEEEauthorblockA{\textit{Berkeley Education Alliance for}\\
\textit{Research in Singapore (BEARS)}\\
Singapore \\
seshucontrol@gmail.com}
}
\maketitle
\begin{abstract}
This investigation aims to study different adaptive fuzzy inference algorithms capable of real-time sequential learning and prediction of time-series data. A brief qualitative description of these algorithms namely meta-cognitive fuzzy inference system (McFIS), sequential adaptive fuzzy inference system (SAFIS) and evolving Takagi-Sugeno (ETS) model provide a comprehensive comparison of their working principle, especially their unique characteristics are discussed. These algorithms are then simulated with dataset collected at one of the academic buildings at Nanyang Technological University, Singapore. The performance are compared by means of the root mean squared error (RMSE) and non-destructive error index (NDEI) of the predicted output. Analysis shows that McFIS shows promising results either with lower RMSE and NDEI or with lower architectural complexity over ETS and SAFIS. Statistical Analysis also reveals the significance of the outcome of these algorithms.  
\end{abstract}


\section{Introduction}
Increasing interest for developing smart energy management system within scientific community has thrived for improvement of short term energy prediction (STEP) algorithms with high accuracy without any computational overload. Model predictive controller based energy management systems are capable of dealing with the uncertainties in energy demand to some extent by taking receding horizon approach. However, receding horizon control requires prediction of future renewable generation and demand (thermal and electrical) beforehand. The short-term prediction helps the energy management system to schedule the energy sources in more cost efficient way avoiding under or over energy generation, plan maintenance work without compromising consumers' comfort \cite{landa2015}. The performance of the algorithms are affected due to several factors. The renewable generation is often subjected to fluctuations induced due to meteorological factors such as irradiation, wind-speed, dust cover etc, which are both intrinsic and extrinsic to the operation of the PV panels. Similarly, both thermal and electrical demand are time-varying parameters that depend on numerous factors which includes type of the day (working day or weekend), month of a year, climatic conditions and so on. These factors are inherently non-linear and time-varying. 

Fuzzy inference systems (FIS) have been evolved successfully for solving different real-time problems due to continuous input-output mapping and interpretation abilities \cite{wu2011continuity,rong2007adaptive,chatterjee2008augmented,uuguz2012adaptive}. In particular, since their advent, neural-fuzzy approaches have become the foremost tool as they inherently assimilate both the learning capability of a neural network and the ability of a FIS to capture and model underlying non-linear characteristics of real-life data with promising accuracies \cite{jang1993anfis}. Kasabov proposed one of the first adaptive neuro-fuzzy inference systems (NFIS) \cite{kasabov2001evolving}, in which rules and parameters are updated by the guidance of a hybrid online supervised/unsupervised learning scheme in response to new ensuing data.  Dynamic evolving neuro-fuzzy inference system (DENFIS)\cite{kasabov2002denfis} uses a clustering method to evolve the rules and update parameters. This network chooses \textit{m}-most significant rules for prediction through the offline clustering technique, which makes DENFIS not suitable for online circumstances. Dynamic fuzzy neural network (D-FNN) \cite{wu2000dynamic,wu2001fast} dynamically adjusts the width of the RBF unit of the TSK-based extended RBF neural network by a hierarchical online self-organized learning depending on the total training data. The absence of total training data limits its usage in offline learning only. A self-constructing neuro-fuzzy inference network (SONFIN) \cite{juang1998online} proposes an input data alignment scheme for clustering and measures a projection-based correlation for evolving rules.

This paper focuses on three of the adaptive FIS that, in true sense, implements sequential learning strategies for rule and parameter, namely meta-cognitive fuzzu inference system (McFIS) \cite{Subramanian2012}, sequential adaptive fuzzy inference system (SAFIS) \cite{rong2006sequential} and evolving Takagi-Sugeno model (ETS) \cite{Angelov2004}. This approaches are well-established in several machine learning problems such as classification, system identification problems \cite{Subramanian2012,angelov2008evolving,rong2006sequential,Subramanian2013}. This investigation analyses and compares the performance of these approaches in terms of their prediction error and architectural complexity for load forecasting in urban buildings and renewable energy generation, where there is a deficiency of storing a large amount of historical data. Also, these dataset shows an ample amount of uncertainties depending on the season, type of the day etc. Apart from this, the behaviour of these approaches is also examined while predicting renewable energy generation with external input to verify the improvement in performance of these FIS.

The rest of the paper is organized as follows. Section II describes the STEP problem definition followed by the working principle of ETS, SAFIS and McFIS. Section III presents the dataset that are used for simulation and case study with parameter settings of these particular algorithms. Section IV concludes the study with a course of potential future implementation of these algorithms in real-time systems.  

\section{Energy Forecasting Problem and Adaptive Neural-Fuzzy Approaches}
This section starts with energy prediction problem definition followed  by a brief yet comprehensive descriptions of the algorithms that are mainly focused for energy demand and renewable generation prediction problems. The algorithms considered here have the capability of online learning and prediction. In case of online learning, the training data are collected and used for parameter update sequentially. The rule base and the parameters are upgraded or modified depending on the strength of the information possessed by the data sample.
\subsection{Energy Forecasting Problem}
STEP addresses the problems of one-hour-ahead to several-day-ahead energy prediction. NFIS learns from a set of training samples given by $\left\{ \left( \mathbf{u^1}, \mathbf{v^1}\right), \dots, \left( \mathbf{u}^k, \mathbf{v}^k\right), \dots \right\}$ where $\mathbf{u}^k = \left[p^k(t),\dots,~p^k(t-\nu+1), r^k(t),\dots,r^k(t-\mu+1) \right]^T \in$ $\mathbb{R}^\nu \times \mathbb{R}^\mu$ is the input vector which consists of the past $\nu$ energy demand time-series sample-points and $\mu$ previous input points to the dynamical system. $\mathbf{v}^k = \left[p^k(t+1),~p^k(t+2),\dots,~p^k(t+\gamma) \right]^T$ $\in \mathbb{R}^m$ is the vector of future responses. $\gamma$ is known as prediction horizon. Forecasting problem can be defined as functional mapping between input and output of a dynamical system $\mathbf{\Phi} : \mathbf{u}^k  \in \mathbb{R}^n \times \mathbb{R}^p \to \mathbf{v}^k \in \mathbb{R}^m$ based on past response of the system. The predicted output of the system is given by
\begin{equation}
\hat{\mathbf{v}}^k = \hat{\mathbf{\Phi}} \left[\mathbf{u}^k, \mathbf{\lambda}\right].
\label{forecast_eqn}
\end{equation}
where, $\mathbf{\lambda}$ is the parameters of the FIS network. The objective is to approximate the function $\mathbf{\Phi[\cdot]}$ such that predicted response $\hat{\mathbf{v}}^k$ is as close as possible to system's actual response $\mathbf{v}^k$.

\subsection{Evolving Takagi-Sugeno Model (ETS)}
ETS proposed by Angelov \textit{et al.} \cite{Angelov2004} uses an on-line clustering technique to gradually evolve Takagi-Sugeno (TS) fuzzy model. It verifies the information content of the data sequentially to update or modify the fuzzy rules. The information content of each data is extracted using a information potential measurement and the spatial proximity of the data samples to the already existing fuzzy rules in the fuzzy sub-space. The algorithmic flow for on-line learning and prediction of ETS is listed as follows.

\begin{enumerate}
\item \textbf{Step 1:} During first iteration, the first data sample is considered as a the focus of the first cluster or rules. This first data sample forms the antecedent part of the first rule using a user-define membership function.
\item \textbf{Step 2:} As the next data sample is considered, the potential of the data sample in the fuzzy rule space is measured recursively using a Cauchy type function by calculating a projection of the distance between the current sample and the previous samples.
\item \textbf{Step 3:} The potential of the cluster centres are updated considering the data samples information in a recursive way using the information from the previous sample points.
\item \textbf{Step 4:} The potential of the new data sample in the fuzzy rule space is compared with the potential of the already existing rule centres. The decision of adding a new rule is made if the potential of the data sample is higher than the potential of the existing cluster centres. Alternatively, the cluster centres' potentials are updated in the next iteration as described in step 3.
\item \textbf{Step 5:} In the penultimate step, The parameters of the consequent part of the rule base are either updated globally using recursive least square (RLS) method or updated locally using weighted RLS technique.
\item \textbf{Step 6:} The final step is to predict the output of the data sample. The iteration is loop again continuous starting from step 2 as the new data sample is collected.
\end{enumerate}
In ETS, as the TS model is evolved in each time step, and the model parameters are modified depending on the gradual change in the cluster centres, the rule base is expected to grow.

\subsection{Sequential Adaptive Fuzzy Inference System (SAFIS)}
SAFIS \cite{rong2006sequential} uses the idea of the \textit{influence of a fuzzy rule} to upgrade the rule base. In statistical sense, the influence of the fuzzy rule is defined as the contribution of that particular fuzzy rule in predicting the overall output. It uses the distance information of the current sample from the existing rules to update the parameters. Only the parameters related to the nearest rules are updated using extended Kalman filter if the measured distance is below a certain threshold. Alternatively, the new rule is added. It also incorporates rule pruning technique considering that the influence of the particular rule is under pre-defined threshold. The usage of the current sample only for updating parameters and upgrading rule base are the reason of achievement of fast computation.

\subsection{Meta-cognitive Neuro-Fuzzy Inference System (McFIS)}
McFIS as proposed by Subramanian \textit{et al.} \cite{Subramanian2012} has been developed based on simple meta-cognition model of Nelson and Narens \cite{nelson1990metamemory}. The working principle of McFIS differs from other adaptive NFIS as discussed follows. The detail parameter update and rule growing and pruning along with architectural desriptions are given in \cite{Subramanian2012}.

\begin{enumerate}
\item The meta-cognitive unit acts as a self-regulatory learning component which controls the learning mechanism of the cognitive component by assessing the current knowledge and identifying the new knowledge based on the state of the cognitive component.

\item Depending on the prediction error knowledge, meta-cognition regulates the learning ability of the main NFIS network with \textit{how-to-learn}, \textit{when-to-learn} and \textit{what-to-learn}, for each samples observed sequentially by the network. Thus, McFIS has the ability to escape over-training.  
\item In order to address the aforementioned learning strategies, McFIS takes three simple actions: {\em{(i)}} remove samples with similar information (\textit{sample deletion}); {\em{(ii)}} grow or prune rules and parameter update depending on the information content of the current sample (\textit{sample learning}); {\em{(iii)}} use less informative samples at a later stage of the learning process to tune the parameters (\textit{sample reserve}). This helps McFIS to avoid over-training but to keep generalization ability.
\end{enumerate}  

\begin{table}[h]
\centering
\caption{1-h-ahead energy demand prediction error comparison}
\begin{tabular}{|c|c|c|c|}
\hline
\textbf{NFIS} & \textbf{RMSE} & \textbf{NDEI} & \textbf{RULES} \\ \hline
eTS & 0.1558 & 0.5948 & 12 \\ \hline
SAFIS & 0.2291 & 0.8660 & 17 \\ \hline
McFIS & \textbf{0.1555} & \textbf{0.5878} & \textbf{5} \\ \hline
\end{tabular}
\end{table}
\begin{table}[h]
\centering
\caption{5-h-ahead energy demand prediction error comparison}
\begin{tabular}{|c|c|c|c|}
\hline
\textbf{NFIS} & \textbf{RMSE} & \textbf{NDEI} & \textbf{RULES} \\ \hline
eTS & 0.2094 & 0.7439 & \textbf{6} \\ \hline
SAFIS & 0.2947 & 1.0310 & 22 \\ \hline
McFIS & \textbf{0.1657} & \textbf{0.5798} & 28 \\ \hline
\end{tabular}
\end{table}
\begin{table}[h]
\centering
\caption{5-min-ahead energy demand prediction error comparison}
\begin{tabular}{|c|c|c|c|}
\hline
\textbf{NFIS} & \textbf{RMSE} & \textbf{NDEI} & \textbf{RULES} \\ \hline
eTS & 0.1019 & 0.5021 & 16 \\ \hline
SAFIS & 0.1174 & 0.5793 & 162 \\ \hline
McFIS & \textbf{0.0964} & \textbf{0.4755} & \textbf{15} \\ \hline
\end{tabular}
\end{table}
\begin{table}[h]
\centering
\caption{1-h-ahead renewable energy prediction error comparison}
\begin{tabular}{|c|c|c|c|}
\hline
\textbf{NFIS} & \textbf{RMSE} & \textbf{NDEI} & \textbf{RULES} \\ \hline
eTS & 0.1334 & 0.6519 & 19 \\ \hline
SAFIS & 0.1496 & 0.7315 & 167 \\ \hline
McFIS & \textbf{0.1328} & \textbf{0.6491} & \textbf{5} \\ \hline
\end{tabular}
\end{table}
\begin{table}[h]
\centering
\caption{5-min-ahead renewable energy prediction error comparison including temperature variation}
\begin{tabular}{|c|c|c|c|}
\hline
\textbf{NFIS} & \textbf{RMSE} & \textbf{NDEI} & \textbf{RULES} \\ \hline
eTS & 0.0909 & 0.5323 & 25 \\ \hline
SAFIS & 0.0976 & 0.5724 & 163 \\ \hline
McFIS & \textbf{0.0867} & \textbf{0.5085} & \textbf{5} \\ \hline
\end{tabular}
\end{table}
\begin{table}[h]
\centering
\caption{1-h-ahead renewable energy prediction error comparison including temperature variation}
\begin{tabular}{|c|c|c|c|}
\hline
\textbf{NFIS} & \textbf{RMSE} & \textbf{NDEI} & \textbf{RULES} \\ \hline
eTS & 0.1365 & 0.7923 & 23 \\ \hline
SAFIS & 0.1239 & 0.7510 & 161 \\ \hline
McFIS & \textbf{0.1221} & \textbf{0.7114} & \textbf{7} \\ \hline
\end{tabular}
\end{table}

\section{Experimental Analysis}
\subsection{Description of the Dataset}
The energy demand data is measured in one of the academic buildings at Nanyang technological university campus, Singapore. The data is collected during the month of May, 2015 with a sampling interval of 1 h over a period of 7 days. Four previous samples are used as the input vector for learning and prediction in all of the algorithms i.e. $\nu=4$. $85$\% of the data points are used for sequential learning of these algorithms. For 1-h and 5-h ahead prediction of energy demand, $\gamma$ is considered to be 1 and 5 respectively.

The renewable energy generation data has been taken during the month of January, 2016 with sampling interval of 5 min. 70\% of the complete dataset is used for training. As similar to the energy data, $\nu=4$ is considered. For 5-min and 1-h ahead prediction of renewable energy, $\gamma$ is set at 1 and 12 respectively.
\begin{figure*}
    \centering
    \begin{subfigure}{.4\linewidth}
    \centering
        \includegraphics[scale=0.45]{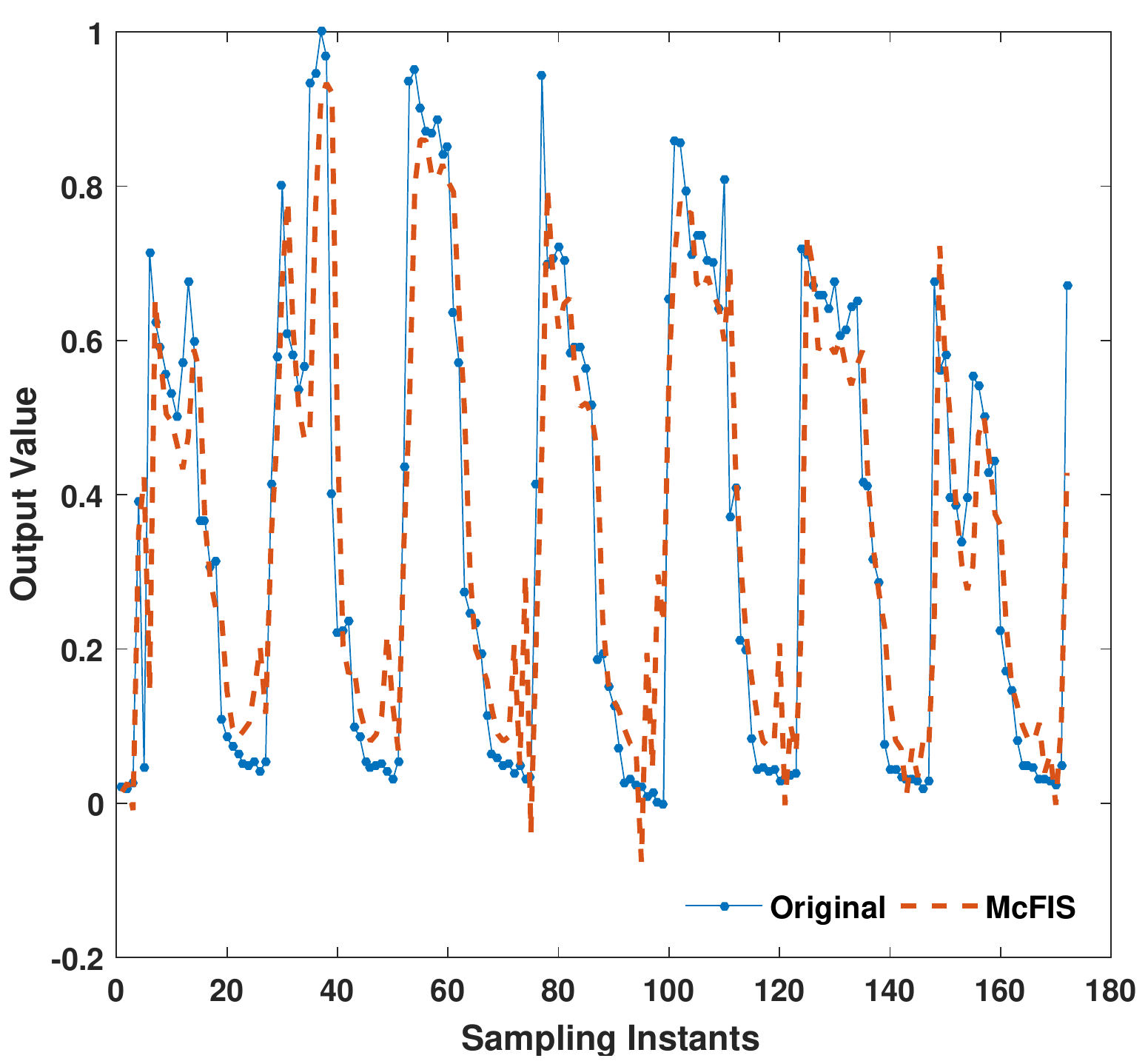}
        \caption{}
        \label{subfig:conv1}
    \end{subfigure}
    \hspace{10pt}
    \begin{subfigure}{.4\linewidth}
    \centering
        \includegraphics[scale=0.45]{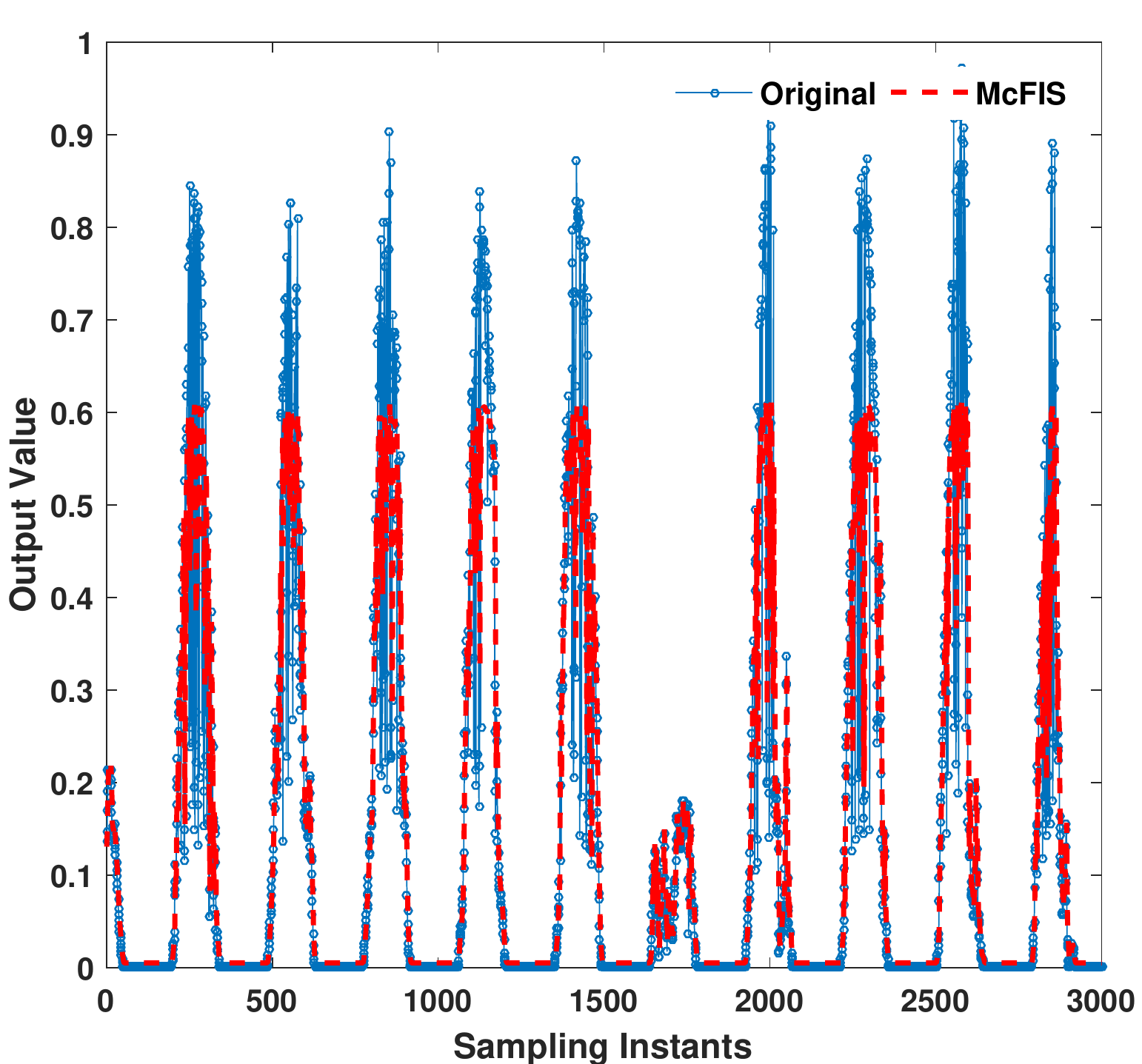}
        \caption{}
        \label{subfig:conv2}
    \end{subfigure}
    \begin{subfigure}{.4\linewidth}
    \centering
        \includegraphics[scale=0.45]{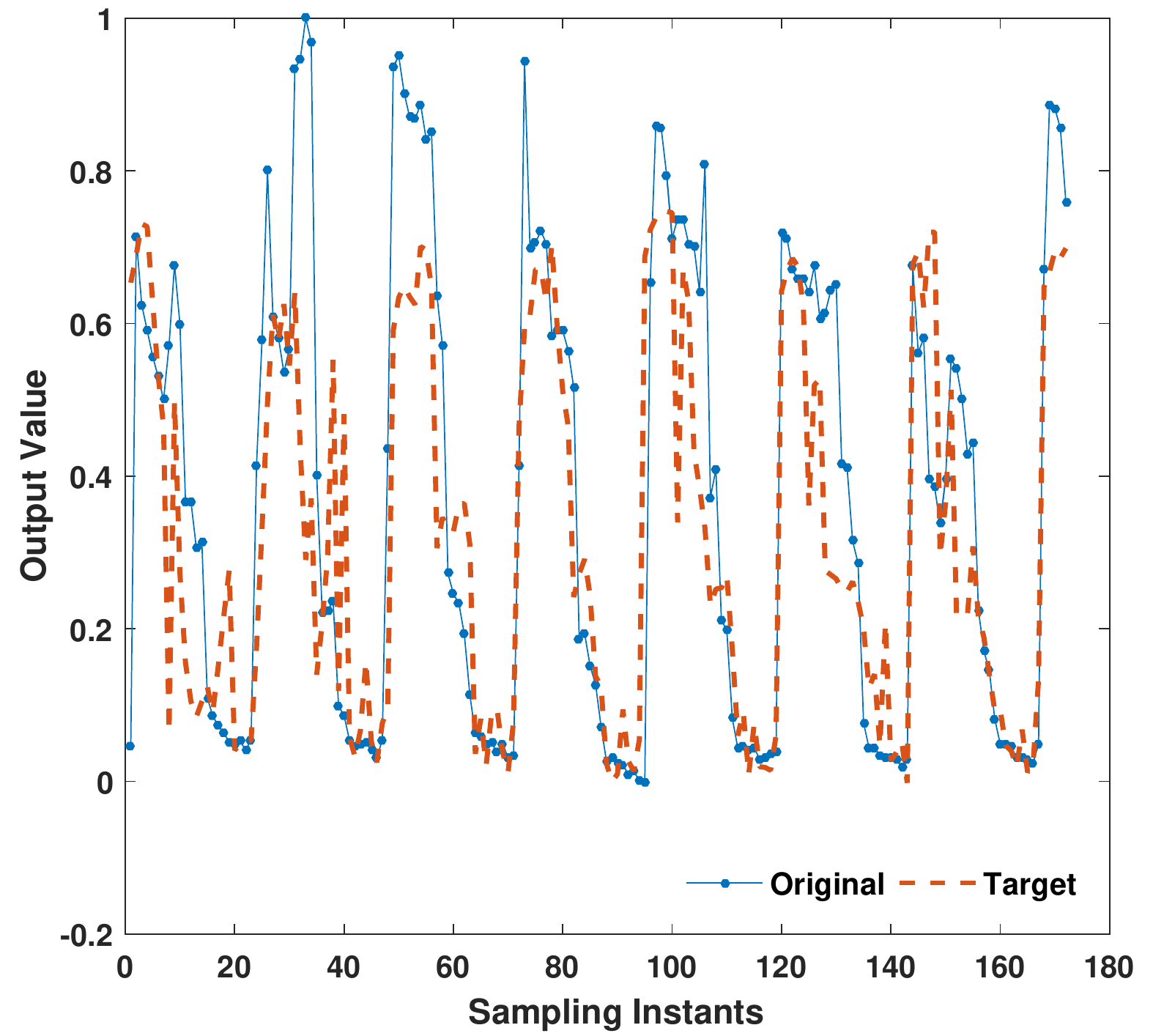}
        \caption{}
        \label{subfig:conv4}
    \end{subfigure}
    \begin{subfigure}{.4\linewidth}
    \centering
        \includegraphics[scale=0.45]{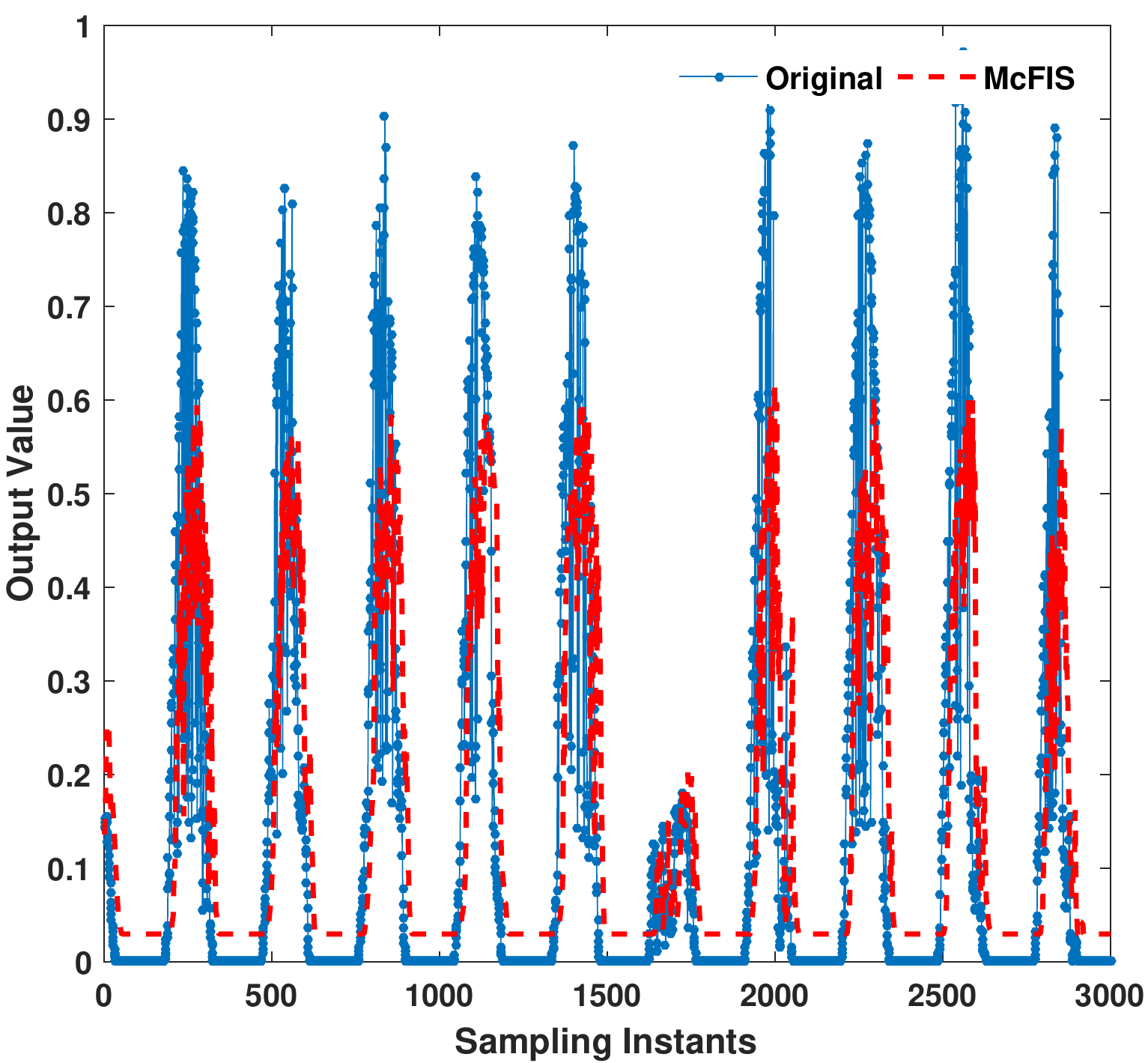}
        \centering
        \caption{}
        \label{subfig:conv5}
    \end{subfigure}
    \caption{Actual vs predicted output data using McFIS. (a) 1-h-ahead energy demand prediction, (b) 5-min-ahead renewable energy prediction, (c) 5-h-ahead energy demand prediction, (d) 1-h-ahead renewable energy prediction.}
    \label{fig:conv1}
\end{figure*}
\subsection{Prediction Performance Analysis}
The prediction errors of these approaches are presented in Table I-IV. The original dataset is normalized for further analysis and prediction. It is to be mentioned that, this analysis only considers time-series prediction, in which any external input to the dynamical system i.e. $u$ is absent. Two error measures termed as root mean squared error (RMSE) and non-destructive error index (NDEI) are used for analysing performance. Besides, the number of fuzzy rules used is also tabulated to reflect the architectural complexity of these inference systems during prediction. 

In case of 1-h-ahead energy demand forecasting as shown in Table I, the prediction accuracies of both McFIS and ETS are comparable. However, McFIS achieves this accuracy using less fuzzy rules, which reduces network architecture complexity. Although McFIS uses a large number of fuzzy rules for 5-h-ahead energy demand forecasting, it reduces the prediction error of ETS and SAFIS by approximately 20\% and 40\% respectively as shown in Table II. In case of renewable energy prediction, McFIS performs significantly well in terms of both RMSE and fuzzy rules over other two approaches as depicted in Table III and IV.

The predicted value of the McFIS is depicted in Figure 1. Figure 1(a) and Figure 1(c) illustrates that McFIS is able to predict the sharp changes in time series dynamics. Similarly it is able to detect and predict the quick changing dynamics of the renewable energy prediction, as shown in Figure 1(b) and 1(d), although it is prominent in Figure 1(d) that, during night hours when there is no output from photo voltaic panel due to absence of solar irradiation, McFIS predicts a non-zero output.

\begin{figure*}
    \centering
    \begin{subfigure}{.3\linewidth}
    \centering
        \includegraphics[scale=0.33]{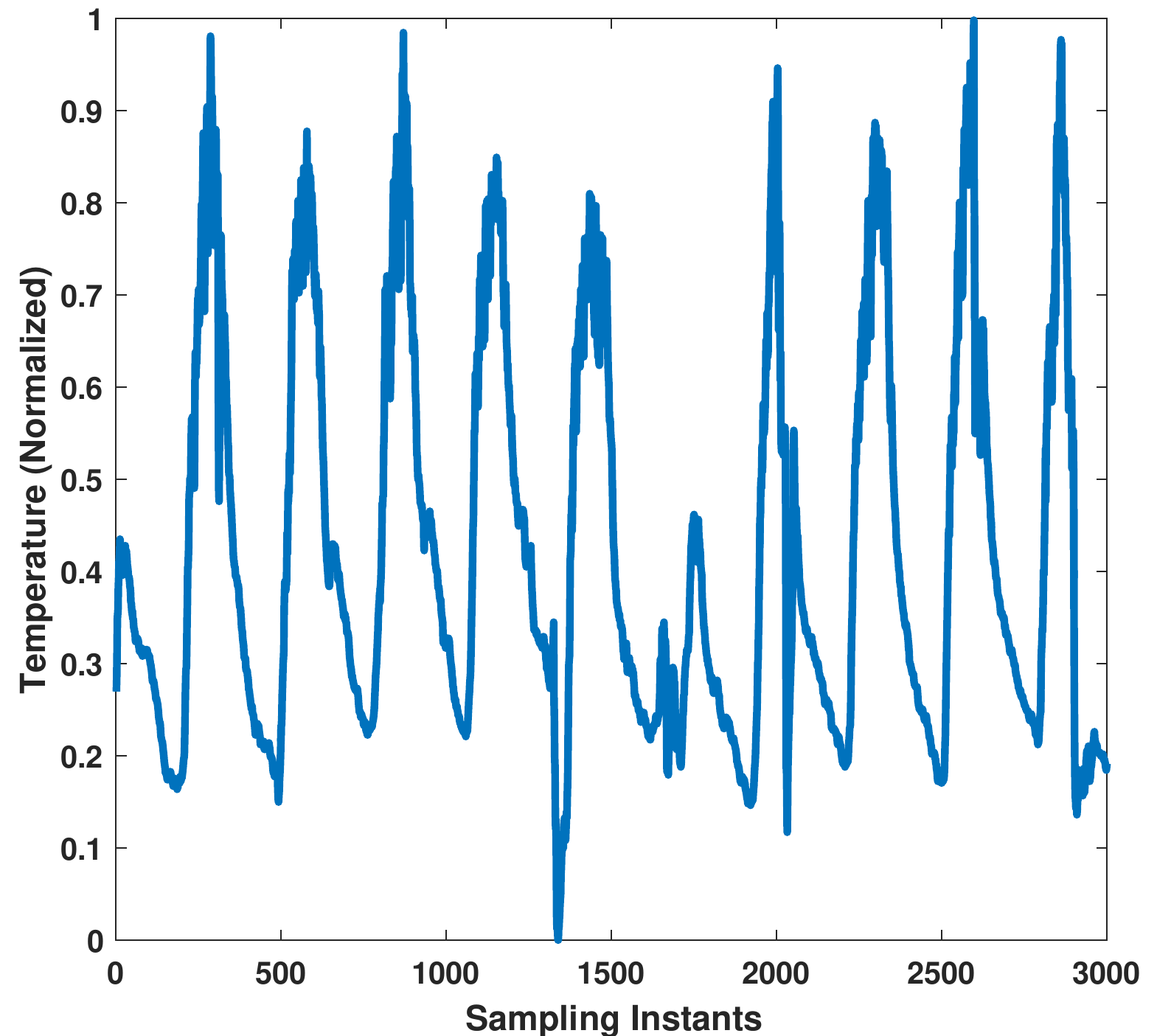}
        \caption{}
        \label{subfig:conv1}
    \end{subfigure}
    \begin{subfigure}{.3\linewidth}
    \centering
        \includegraphics[scale=0.33]{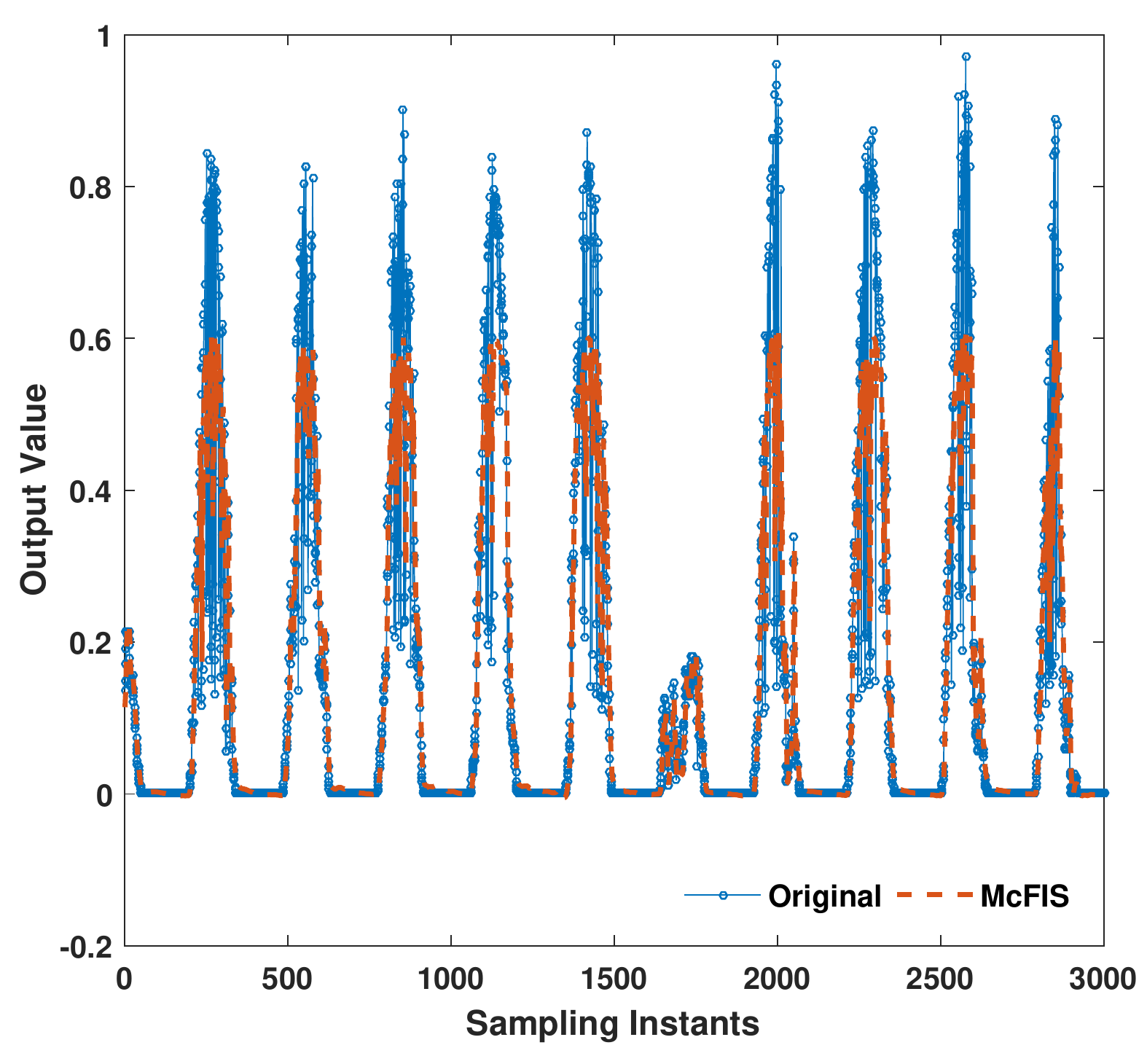}
        \caption{}
        \label{subfig:conv2}
    \end{subfigure}
    \begin{subfigure}{.3\linewidth}
    \centering
        \includegraphics[scale=0.33]{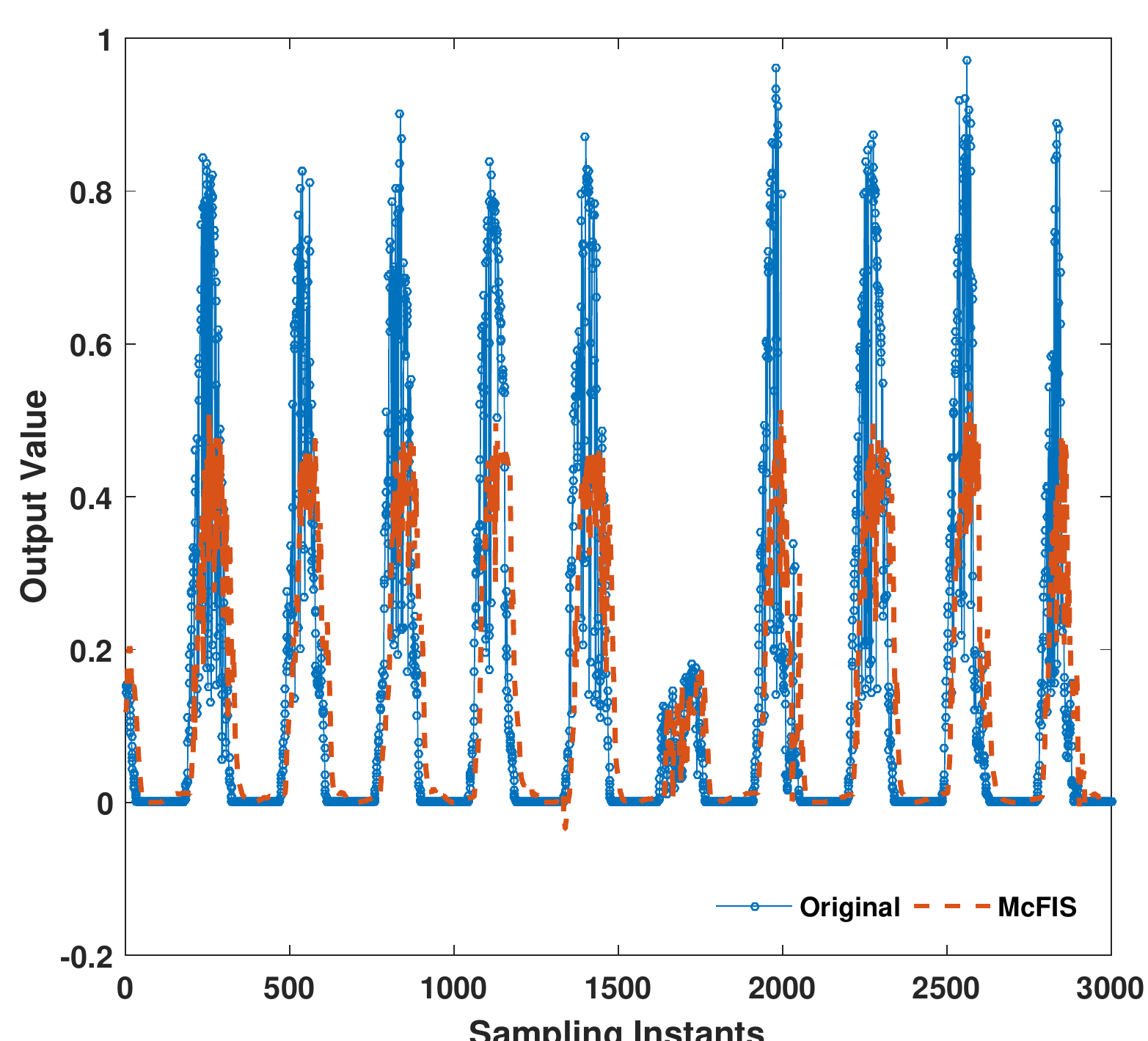}
        \caption{}
        \label{subfig:conv1}
    \end{subfigure}
    \caption{Renewable energy prediction with temperature as input signal using McFIS. (a) Temperature variation, (b) 5-min-ahead renewable energy prediction, (c) 1-h-ahead renewable energy prediction}
    \label{fig:conv1}
\end{figure*}
\subsection{Renewable Energy Prediction with Temperature Variation}
This investigation also considers temperature as an input feature to the dynamical system to predict the renewable energy generation. Only current instant temperature is considered, i.e. $\mu=1$. The temperature variation as measured at a sampling period of 5 min during January 2016 over a period of 10 days is shown in Figure 2. Table V and VI tabulates the prediction error and number of fuzzy rules used. One can observe that although eTS and SAFIS fails to achieve a significant improvement, McFIS certainly improves the prediction error for 1-h-ahead forecasting and architectural complexity in both 5-min and 1-h-ahead prediction problems. Also it can be observed from Figure 2(c) that incorporating temperature as input variable helps to increase the prediction accuracy of McFIS during no-energy-generation hours.
\subsection{Rank-based Statistical Comparison}
In order to find the statistical significance of the outcome of McFIS algorithm, Friedman's rank test is realized. For ease of understanding, we denote the problems as follows.

\begin{enumerate}
\item[] \textit{F$_1$}: 1-h-ahead energy demand prediction,
\item[] \textit{F$_2$}: 5-h-ahead energy demand prediction,
\item[] \textit{F$_3$}: 5-min-ahead renewable energy prediction,
\item[] \textit{F$_4$}: 1-h-ahead renewable energy prediction,
\item[] \textit{F$_5$}: 5-min-ahead renewable energy prediction with temperature as input,
\item[] \textit{F$_6$}: 1-h-ahead renewable energy prediction with temperature as input.
\end{enumerate}

\begin{table}[]
\centering
\caption{Average ranking of the algorithms}
\begin{tabular}{l|c|c|c}
\hline
\textbf{Probelms} & \textbf{ETS} & \textbf{SAFIS} & \textbf{McFIS} \\ \hline
\textbf{F1} & 2 & 3 & 1 \\ \hline
\textbf{F2} & 2 & 3 & 1 \\ \hline
\textbf{F3} & 2 & 3 & 1 \\ \hline
\textbf{F4} & 2 & 3 & 1 \\ \hline
\textbf{F5} & 2 & 3 & 1 \\ \hline
\textbf{F6} & 3 & 2 & 1 \\ \hline
\textbf{Avg Rank} & 2.17 & 2.83 & \textbf{1} \\ \hline
\end{tabular}
\end{table}

\begin{table}[]
\centering
\caption{Average rank difference and statistics}
\begin{tabular}{c|c}
\hline
Algorithm & \begin{tabular}[c]{@{}l@{}}Difference between average \\ ranking w.r.t McFIS\end{tabular} \\ \hline
ETS & 1.17 \\ \hline
SAFIS & 1.83 \\ \hline
Critical difference (Bonferroni-Dunn) & \begin{tabular}[c]{@{}l@{}}1.3 ($\alpha$ = 0.05) , \\ 1.13 ($\alpha$ = 0.01)\end{tabular} \\ \hline
\end{tabular}
\end{table}

Table VII shows the rank, average rank of the these algorithms considering RMSE as the performance metric. Friedman test initially assumes that all the algorithms are performing similar. Depending on the average ranking, the test computes a $Q$ value. In this study, the Q value is found as 10.3068 which is greater than the Friedman statistical value atconfidence level of 95\% and 99\% which are 7 and 9 respectively for a problem with 3 treatments and 6 blocks problem \cite{demvsar2006}. As a result, the null hypothesis can be rejected and it can be concluded that the average ranking is statistically significant. In our case, as the number of hypothesis and number of dataset are low, a pairwise post-hoc Bonferroni-Dunn test will verify the statistical significance of the predictive algorithms under investigation which is recorded in Table VIII. This particular test states that the performance of a particular algorithm has statistical impact if the difference between the average rank is greater than a critical difference with some confidence \cite{demvsar2006}. Table VIII signifies that the performance of McFIS is better than that of SAFIS and eTS with a 99\% confidence level.

\section{Conclusion}
This investigation studies the performance of some of the well-established adaptive FIS in prediction short term energy demand and renewable generation. From the analysis it is found that McFIS has the ability to predict data with the promising accuracy by keeping the number of fuzzy rules as low as possible. It has also been observed that presence of external relevant parameters e.g. temperature variation in renewable generation not only improves computational requirement in McFIS but also shows a promising direction towards improving the prediction error. The low computational requirement as well as high prediction accuracy makes McFIS more suitable forecasting tool for STEP in urban buildings.   
\bibliographystyle{IEEEtran}
\bibliography{IEEESSCIForecasting}
\end{document}